\documentclass[journal]{IEEEtran}
\usepackage{amsmath,amsfonts}
\usepackage{algorithmic}
\usepackage{algorithm}
\usepackage{array}
\usepackage[caption=false,font=normalsize,labelfont=sf,textfont=sf]{subfig}
\usepackage{textcomp}
\usepackage{stfloats}
\usepackage{url}
\usepackage{verbatim}
\usepackage{graphicx}
\usepackage{multirow}
\usepackage{orcidlink} 
\usepackage{marvosym}

\usepackage{cite}
\begin{document}

\title{FitPro: A Zero-Shot Framework for Interactive Text-based Pedestrian Retrieval in Open World}

\author{
  Zengli Luo\orcidlink{0009-0002-3032-0956},~\IEEEmembership{Member,~IEEE},
  Canlong Zhang\orcidlink{0000-0003-4375-1405},~\IEEEmembership{Member,~IEEE},
  Xiaochun Lu, 
  and Zhixin Li\orcidlink{0000-0002-5313-6134}
  \thanks{This work was supported by the National Natural Science Foundation of China under Grant 62266009; in part by the Guangxi Science and Technology Program of Guangxi Engineering Research Center of Educational Intelligent Technology under Grant AB25069418 and Grant 2018GXNSFDA281009. (\textit{Corresponding author: Canlong Zhang}).}
  \thanks{Zengli Luo is with the Key Lab of Education Blockchain and Intelligent Technology, Ministry of Education, Guangxi Normal University, Guilin, China, and also with the School of Basic Medical Sciences, Guangzhou University of Chinese Medicine, Guangzhou, China.}
  \thanks{Canlong Zhang (\textsuperscript{\Letter}) is with the Key Lab of Education Blockchain and Intelligent Technology, Ministry of Education, Guangxi Normal University, Guilin, China, and also with the Guangxi Key Lab of Multi-source Information Mining \& Security, Guangxi Normal University, Guilin, China (e-mail: clzhang@gxnu.edu.cn).}
  \thanks{Xiaochun Lu and Zhixin Li are with the Key Lab of Education Blockchain and Intelligent Technology, Ministry of Education, Guangxi Normal University, Guilin, China, and also with the Guangxi Key Lab of Multi-source Information Mining \& Security, Guangxi Normal University, Guilin, China.}
  \thanks{The code is available at \url{https://github.com/lilo4096/FitPro-Interactive-Person-Retrieval}.}
}
\markboth{Journal of \LaTeX\ Class Files,~Vol.~14, No.~8, August~2025}%
{Luo \MakeLowercase{\textit{et al.}}: FitPro: A Zero-Shot Framework for Interactive Text-based Pedestrian Retrieval in Open World}

\maketitle

\begin{abstract}
  Text-based Pedestrian Retrieval (TPR) deals with retrieving specific target pedestrians in visual scenes according to natural language descriptions. 
  Although existing methods have achieved progress under constrained settings, interactive retrieval in the open-world scenario still suffers from limited model generalization and insufficient semantic understanding. 
  To address these challenges, we propose FitPro, an open-world interactive zero-shot TPR framework with enhanced semantic comprehension and cross-scene adaptability.  
  FitPro has three innovative components: Feature Contrastive Decoding (FCD), Incremental Semantic Mining (ISM), and Query-aware Hierarchical Retrieval (QHR). 
  The FCD integrates prompt-guided contrastive decoding to generate high-quality structured pedestrian descriptions from denoised images, effectively alleviating semantic drift in zero-shot scenarios. 
  The ISM constructs holistic pedestrian representations from multi-view observations to achieve global semantic modeling in multi-turn interactions, 
  thereby improving robustness against viewpoint shifts and fine-grained variations in descriptions. 
  The QHR dynamically optimizes the retrieval pipeline according to query types, enabling efficient adaptation to multi-modal and multi-view inputs. 
  Extensive experiments on five public datasets and two evaluation protocols demonstrate that FitPro significantly overcomes the generalization limitations and semantic modeling constraints of existing methods in interactive retrieval, paving the way for practical deployment. 
\end{abstract}

\section{Introduction}
Text-based Person Re-identification (TPR) has emerged as a key technology in various open-world applications~\cite{TPRSurvey, TPRSurvey2}. 
Conventionally, TPR is addressed via supervised learning frameworks relying on large-scale labeled datasets for deep model training.
However, collecting and annotating such data incurs exceedingly time-consuming,and models trained on a single scene often fail to generalize to new environments due to domain discrepancies~\cite{ConventionalLimitations, lowprecisionGPULLM}. 
Besides, dynamic real-world surveillance requires prompt responses to sudden queries, whereas the high cost of repeated data collection and annotation limits system flexibility~\cite{DynamicQuery}. 
These challenges call for frameworks with strong generalization and easy deployment, particularly in multi-source surveillance systems where scene-specific training remains difficult.
In this context, retrieval models with zero-shot adaptability are urgently required~\cite{ReviewHumanCentric}.

\begin{figure}[!t]
  \centering
  \includegraphics[width=\columnwidth]{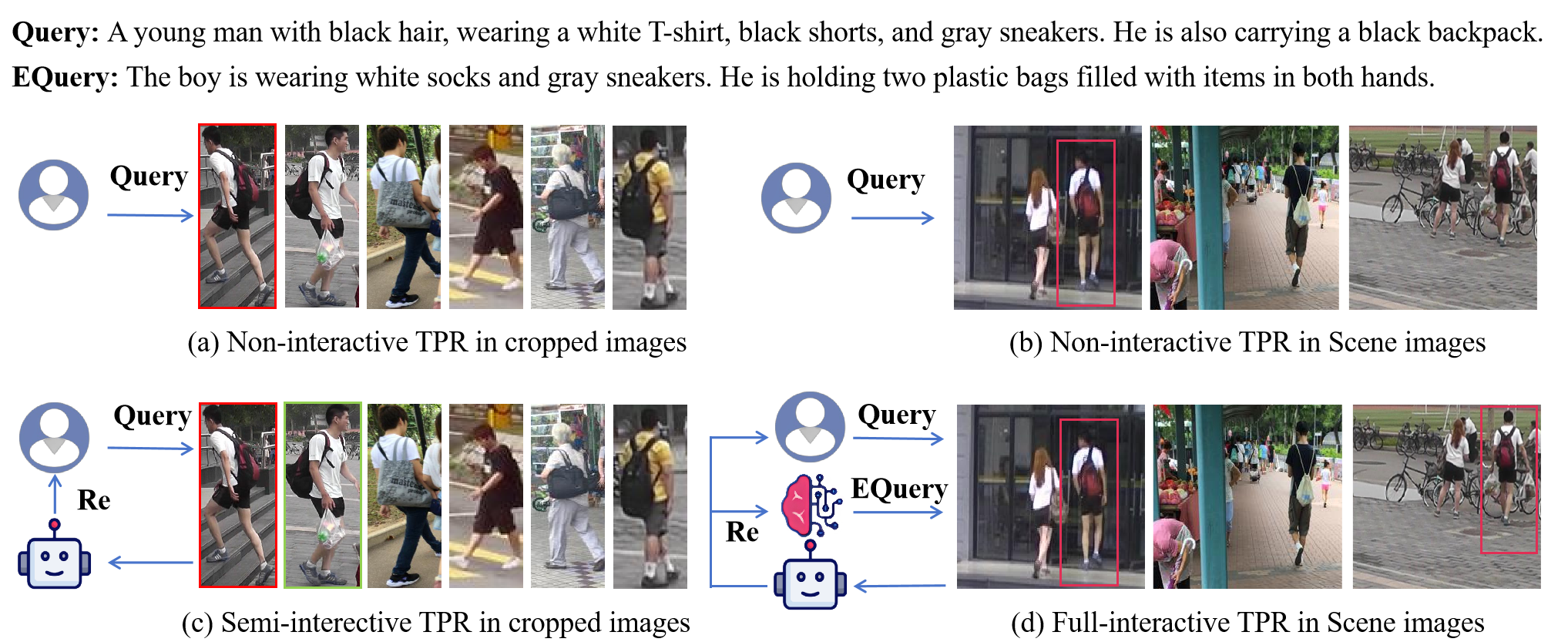}
  \caption{Comparison of four TPR paradigms.}
	\label{fig1}
\end{figure}

Existing TPR methods have primarily focused on cropped-image retrieval, where the target pedestrian is retrieved from a gallery based on textual descriptions (Fig.~\ref{fig1}(a))~\cite{IRRA}, \cite{AttributeReid}, \cite{FLAN}, \cite{LLaVA-ReID}. 
Although such methods have made significant progress, they often involve pre-detected bounding boxes that require further post-processing, 
which restricts their effectiveness in handling full-scene or open-world scenarios. 
Besides, variations in camera angle, occlusion, and environmental factors often lead to degraded matching performance, 
similar to challenges observed in related domains such as 3D human pose estimation and visible-infrared person re-identification~\cite{3DPose, VIReid}.

To alleviate these issues, several studies, including our previous work\cite{SDRPN}, \cite{UPD},\cite{MACA}, have proposed end-to-end, scene-level TPR frameworks (Fig.~\ref{fig1}(b)), 
which jointly perform detection, identification, and image-text matching to directly retrieve pedestrians from entire images. 
While these approaches eliminate the need for external bounding-box detectors, they are still constrained by static attribute modeling, 
which limits their robustness to complex viewpoint variations and temporal changes, potentially leading to mismatches.

Recently, semi-interactive frameworks (Fig.~\ref{fig1}(c)) have been explored to balance efficiency and flexibility~\cite{Instruct-reid},\cite{CIM},\cite{ChatReID},\cite{PromptTPR},
where user feedback is incrementally integrated through memory mechanisms or progressive refinement strategies.
Although these methods enhance the system's ability to capture fine-grained semantics, they still face several key challenges, 
including discrete semantic representations, static viewpoint modeling, and precision-oriented retrieval strategies that often compromise recall in open-world settings~\cite{Hulk, MultiQA}.  

Furthermore, in real world interactive settings, users rarely provide full descriptions at once~\cite{MultiQA}; 
instead, queries evolve progressively—for example, a pedestrian may initially be described as "carrying a black backpack" and later refined to "a red-and-black backpack" (Fig.~\ref{fig1}(d)). 
Existing methods still lack mechanisms for continuous semantic modeling and contextual integration, 
susceptible to retrieval inaccuracies retrieval failures when key details are missing or ambiguously expressed. 
Recent studies~\cite{CognitiveBiases,humancentric} also emphasized the importance of adaptability and human-centric design,
yet these aspects remain underexplored in current TPR research.

To systematically mitigate the above issues, this work aims to develop a TPR framework that enables \textbf{zero-shot cross-scene retrieval}, supports \textbf{progressive multi-turn interaction}, 
and adaptively balances \textbf{recall-precision trade-offs} according to query completeness. 
To this end, we propose \textit{FitPro}, an interactive zero-shot TPR framework for open-world scenarios (Fig.~\ref{fig1}(d)). 
FitPro achieves cross-scene adaptability through architectural optimization and enhances robustness via multi-turn semantic modeling and efficient retrieval strategies. 
Specifically, FitPro consists of three key modules: 
(i) a Feature Contrastive Decoding (FCD) module that generates robust structured text-image representations from denoised images by aligning semantics, 
alleviating matching drift under viewpoint changes or incomplete descriptions; 
(ii) an Incremental Semantic Mining (ISM) module that integrates multi-view and multi-turn semantics through graph-based representations, 
dynamically maintaining the semantic trajectory of user queries to better handle temporal variations and progressive interactions; and 
(iii) a Query-aware Hierarchical Retrieval (QHR) module that adaptively optimizes the retrieval pipeline according to query types, 
improving efficiency and scalability in open-world scenarios.

In summary, this paper presents a unified framework for zero-shot adaptation and interactive semantic understanding for open-world TPR,  
thereby enhancing generalization, semantic comprehension, and interaction adaptability. The key contributions are:
\begin{enumerate}
  \item We propose an \textbf{interactive zero-shot TPR framework}, \textit{FitPro}, enabling cross-scene generalization without additional training on target domains, addressing the deployment challenges and limited adaptability of existing methods.
  \item We design three key mechanisms to improve robustness in open-world scenarios: the \textbf{FCD} module for stable text-image alignment, the \textbf{ISM} module for structured modeling of semantic evolution, and the \textbf{QHR} module for synergistic optimization of recall and precision.  
  \item We conduct extensive experiments on five benchmark datasets under two evaluation protocols, showing that \textit{FitPro} consistently \textbf{outperforms SOTA methods} in various zero-shot settings, validating its strong generalization capability and practical applicability. 
\end{enumerate}

\section{Related Works}\label{sec2}
\subsection{Text-Based Person Retrieval in the Closed World}  
TPR was initially explored under closed-world assumptions.
Many representative methods~\cite{CUHK}, \cite{ICFG}, \cite{RSTP}, \cite{traditionalTPR} adopt supervised learning frameworks,
where natural language descriptions retrieve pedestrians from cropped images via a joint embedding space.
Subsequent methods~\cite{RASA}, \cite{MARS}, \cite{RDE} improve alignment through attention mechanisms or region-level semantic enhancement,
while some~\cite{IRRA}, \cite{APTM} introduced guided decoders or cross-modal contrastive learning to further improve multi-modal consistency. 
These approaches substantially advance cross-modal representation learning within predefined domains.
However, reliance on large-scale labeled datasets can still limit scalability, 
and models trained within a single scene may face performance challenges when deployed across diverse environments\cite{ConventionalLimitations}, \cite{lowprecisionGPULLM}, \cite{DynamicQuery}.
Furthermore, closed-world frameworks are often limited to fixed identity sets, hindering their adaptability to new scenes with unseen pedestrians.
These challenges motivate the exploration of more flexible paradigms capable of operating on full images and generalizing beyond the training domain.

\subsection{Text-Based Person Retrieval in the Open World}
Despite progress in closed-world settings, real-world TPR often encounters identities unseen during training, 
requiring retrieval methods with open-world capability~\cite{Ov-dquo}, \cite{OVDETR}, \cite{declip},\cite{ ovllm}. 
To address this, several works~\cite{SDRPN}, \cite{UPD}, \cite{MACA} develop end-to-end scene-level systems that jointly perform detection, identification, and vision-language matching on full images. 
Although these approaches alleviate the dependence on pre-cropped inputs, some involve complex detection and post-processing procedures like multi-stage cropping or non-maximum suppression\cite{objectDetection},\cite{multiStage},\cite{siamca},
which may increase computational overhead and complicate deployment in diverse full-scene scenarios with occlusion and viewpoint variations.
Moreover, most systems are trained within predefined category spaces, limiting their ability to handle unseen identities or dynamically evolving scenes. 
While unknown samples can be detected, adaptability to evolving identities and changing user queries remains limited. 
Recently, although interactive methods~\cite{InteractiveMultimediaRetrieval} have made progress in the field of image-text retrieval their applicability in TPR scenarios remains limited. 
These challenges highlight the need for adaptive TPR frameworks that can incorporate evolving semantics, handle viewpoint variations, and remain effective under open-world dynamics.

\subsection{Zero-Shot Vision-Language Retrieval and Its Challenges in TPR}
With the rapid development of large-scale vision-language pretraining, zero-shot vision-language retrieval~\cite{dcdl},\cite{evdclip} has emerged as a promising paradigm for improving model generalization. 
Typical approaches are trained on large-scale image-text pairs to learn cross-modal alignment, thereby enabling retrieval without task-specific supervision~\cite{TransTPS}. 
Despite these advances, directly extending such general zero-shot retrieval models\cite{MatchingtoGeneration} to TPR remains non-trivial. 
First, these models are typically optimized for global semantic matching and may lack the fine-grained attribute sensitivity required for pedestrian retrieval. 
Second, their single-turn retrieval mechanisms are often insufficient to incorporate user-provided incremental information in interactive scenarios. 
Recent efforts~\cite{CIM},\cite{ChatReID} explore memory-based and multi-turn interaction strategies to enhance adaptability.
However, effectively integrating evolving semantics across turns remains challenging, particularly for maintaining global contextual consistency over time.
Accordingly, developing a zero-shot TPR framework that supports fine-grained matching and robust multi-turn semantic integration is still an open problem.
\begin{figure*}
	\centering
	\includegraphics[width=6in]{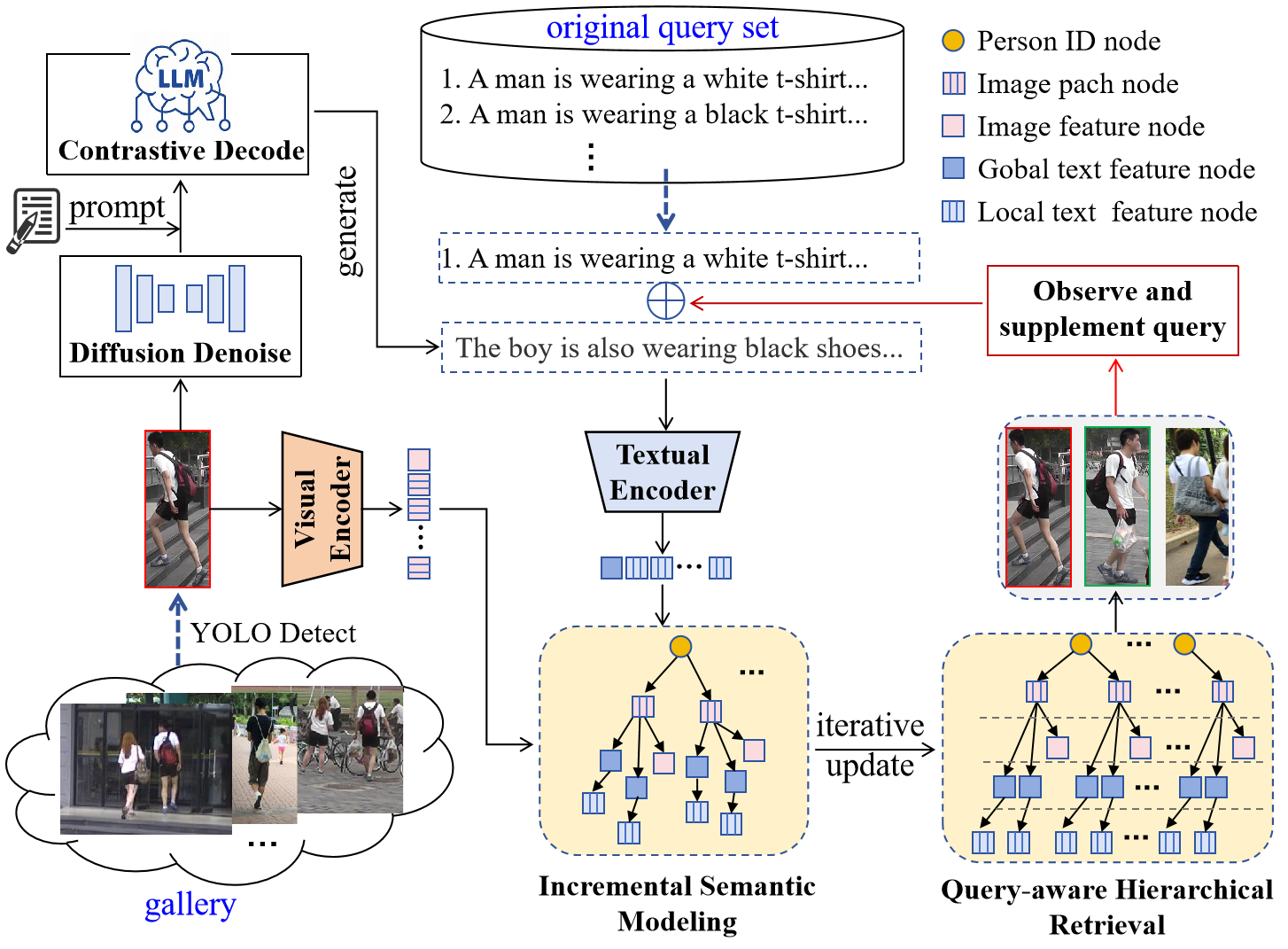}
	\caption{The proposed FitPro framework for text-based zero-shot interactive person retrieval in open scenarios.}
	\label{fig2}
\end{figure*}
\section{Methods}\label{sec3}
\subsection{Overview}
Fig.~\ref{fig2} illustrates the proposed FitPro framework for fully interactive text-based person retrieval in open-world scenarios. 
The framework consists of three key modules: FCD, ISM, and QHR.In contrast to traditional static text-image matching methods, 
FitPro incorporates a user-driven interactive mechanism, enabling the system to dynamically aggregate and update retrieval information across multiple perspectives. 
This interactive design strengthens both the consistency and precision of cross-modal alignment and naturally supports the iterative process described below. 

In round $0$, given an initial query description $Q_{0}$, the FCD module denoises the detected pedestrian regions $I$ by YOLO threshold 
and produces a high-quality pedestrian image $I_{opt}$ along with textual semantics $D={q_{en}, y_{ori}}$ that are consistent with the visual structure.
The ISM module then initializes the pedestrian semantic graph database as $G^{(0)}={q_{en}}$ by incrementally updating nodes to capture the evolving semantics.
Subsequently, the QHR module performs graph-based hierarchical retrieval, yielding the first candidate set $A_{0}$. 

From the first round ($r \geq 1$), the system incorporates the user's additional feedback or supplementary attributes $A_{r-1}$,
updates the semantic query representation, and formulates the revised query $C_{r} = (T_{0}, Q_{0}, A_{0}, \ldots, A_{r})$, 
which is further expanded into a semantic graph $G^{(r)}$, including new entities and relations. 
These enriched semantics are then jointly encoded to ensure coverage of fine-grained attributes and contextual information,
while the QHR module performs hierarchical retrieval with respect to the updated query $C_{r}$, generating a refined candidate set $A_{r}$.

Through this iterative, multiturn, and progressively evolving interaction mechanism, FitPro establishes a closed-loop retrieval paradigm that integrates user feedback and dynamic semantic expansion, 
thereby effectively enhancing robustness and generalization in open-world pedestrian retrieval scenarios.

\subsection{Feature Contrastive Decoding}\label{FCD}
\begin{figure}[!t]
  \centering
  \includegraphics[width=\columnwidth]{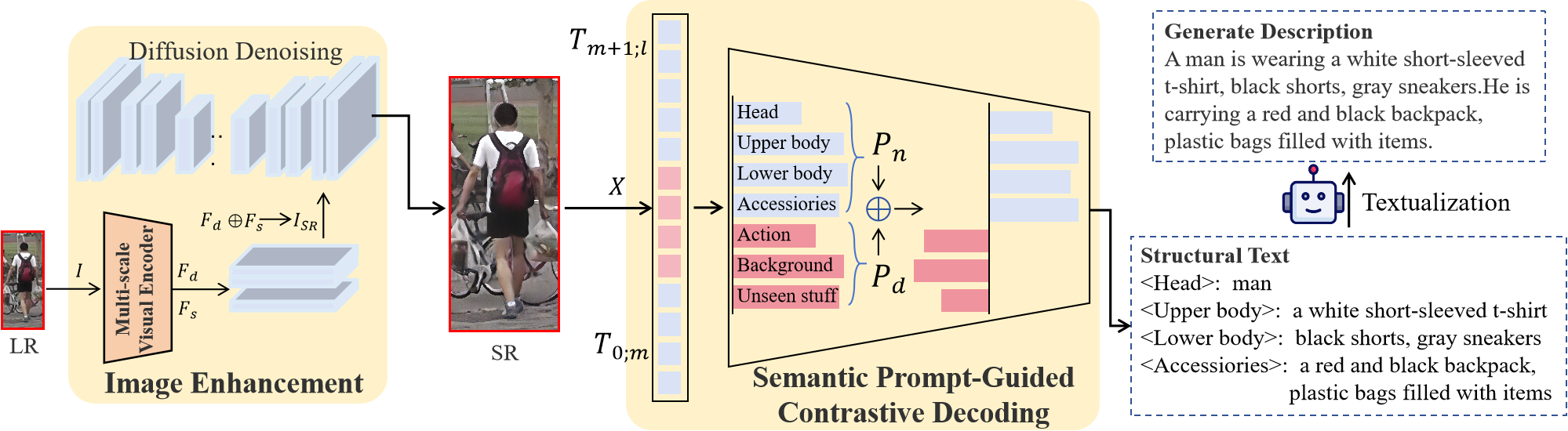}
	\caption{The proposed Feature Contrastive Decoding (FCD) module.}
	\label{fig3}
\end{figure}

To mitigate the impact of pedestrian motion in dynamic scenes, the FCD module introduces a structure-aware diffusion strategy~\cite{ID-Blau}.
By guiding the model to focus on local structures and texture information within pedestrian images (Fig.\ref{fig3}(a)), 
this strategy mitigates the interference of redundant background regions in visual representations. 
The encoded image tokens in Fig.~\ref{fig3}(b) are subsequently input into a large-scale multimodal language model (MLLM). 
Considering the sensitivity of MLLMs to hallucinations~\cite{Evaluating},\cite{MLLMICLR}, \cite{ICD}, \cite{CDECCV}, 
we extend previous contrastive decoding approaches~\cite{IMCCD} by introducing structural priors to guide the decoding process. 
This mechanism suppresses visual illusions and enforces consistency between structural priors and textual semantics, thereby producing more accurate textual descriptions. 
Given an image dataset $I=\{I_{1}, I_{2}, \ldots, I_{N}\}$, where each raw pedestrian image $I_{i}\in\mathbb{R}^{H\times W \times C}$, 
the FCD module first extracts visual features $F_{d}$ and shallow features $F_{0}$ via a visual encoder. 
These features are fused through a convolutional and upsampling fusion layer $H_{rec}$ to reconstruct a high-resolution image $I_{sr}\in\mathbb{R}^{H\times W\times C_{n}}$:
\begin{equation}
    I_{sr} = H_{rec}(F_{0} \oplus F_{d}),
\end{equation}
where $H_{rec}$ denotes the convolution and upsampling fusion layer, and the upsampling factor is empirically set to 4. 

Different from conventional super-resolution approaches that rely on multiscale feature pyramids~\cite{MAN,SCPSN}, the FCD module adopts structural priors as auxiliary constraints to enhance local details and semantic fidelity. 
More precisely, a predefined structural prior map C~\cite{ID-Blau} is introduced as a conditional input for diffusion generation, guiding the model to focus on the reconstruction of edges and high-frequency texture regions. 
This mechanism not only effectively preserves distinctive regional features within the image but also alleviates the over-smoothing issue commonly observed in diffusion-based generation. 
Thereby improves the stability and accuracy of downstream structured semantic generation. 
The diffusion process is implemented using DDIM~\cite{DDIM}, with the iterative procedure formulated as follows:
\begin{equation}
    B_{t-1}=\sqrt{\alpha_{t-1}}\left(B_{t}-\frac{\sqrt{1-\alpha_{t}}}{\sqrt{\alpha_{t}}} \cdot \epsilon_{\theta}\left(B_{t}, I_{sr}, C, t\right)\right)
\end{equation}
Specifically, let $\alpha_{t}\in(0,1)$ denote the diffusion control parameter, and $\epsilon_{\theta}$ be the noise predicted by a pre-trained lightweight denoising UNet~\cite{ID-Blau}. 
The intermediate result at timestep $t$ is denoted as $B_{t}$. The final restored pedestrian image $B_{0}$, obtained after the first denoising stage and the second upsampling stage, serves as the denoised image $I_{opt}$ of the FCD module.

The denoised pedestrian image $I_{opt}$ is then fed into the visual encoder to extract $n$ patch features $\mathbf{X}=\{x_{i}\}_{i=1}^{n}$, 
which are further projected into the input space of the language model as token representations $\mathbf{X}_{tok} = \text{Proj}(\mathbf{X})$.
To mitigate visual hallucinations in multimodal large language models and to strengthen the alignment between textual semantics and structural image content, 
we design a dynamic prompting template $\mathbf{T}$, which consists of a system-level instruction template $T_{sys}$ and an object-level structural template $T_{obj}$. 

In contrast to existing parameterized prompting methods~\cite{PromptTPR},\cite{2024evolving},\cite{PromptPS},\cite{APromptTPR}, the proposed structural semantic template not only constructs the input sequence but also continuously regulates the attention distribution during the generation stage(Fig.~\ref{fig3}(b)). 
This guidance enables the model to focus on discriminative key regions, alleviates semantic drift, and improves the fidelity and structural coherence of generated descriptions.
More concretely, during decoding, a contrastive decoding strategy is introduced to explicitly suppress task-irrelevant regions (e.g., background or dynamic objects), 
thereby enhancing the discriminability and precision of the generated descriptions. 

The final input sequence to the language model is formulated as $S_{in} = [T_{sys}, \mathbf{X}_{tok}, T_{obj}]$, which is processed through step-wise decoding to generate structured pedestrian descriptions $Y={Y_t}$.
The generated description in Fig.~\ref{fig3}(c) $Y$ explicitly covers structured components such as head, upper body, lower body, and accessories, enabling fine-grained external characterization of the target pedestrian. 
These structured descriptions provide a reliable and consistent semantic basis for subsequent multi-turn interactive retrieval.

\subsection{Incremental Semantic Mining}\label{ISM}
\textbf{False Negatives in Single-Round Matching:} Prior work~\cite{RASA} has shown that variations in viewpoint and occlusion during training may enlarge the semantic discrepancy between positive and negative pairs, 
thereby inducing semantic inconsistency between image and text. 
Although probabilistic learning approaches~\cite{MARS} have been explored to mitigate this issue, we observe that such methods still suffer from systematic limitations when directly applied to open-world scenarios. 
Specifically, single-round retrieval results are highly sensitive to transient changes in pedestrian states (e.g., occlusion, illumination), which often introduce false negatives into the results. 
As illustrated in Fig.~\ref{fig1}(d), a single query may only capture the pedestrian from a particular moment and viewpoint, while in real-world scenes, the same pedestrian may exhibit diverse visual appearances and corresponding textual descriptions. 
\begin{figure}[!t]
  \centering
  \includegraphics[width=\columnwidth]{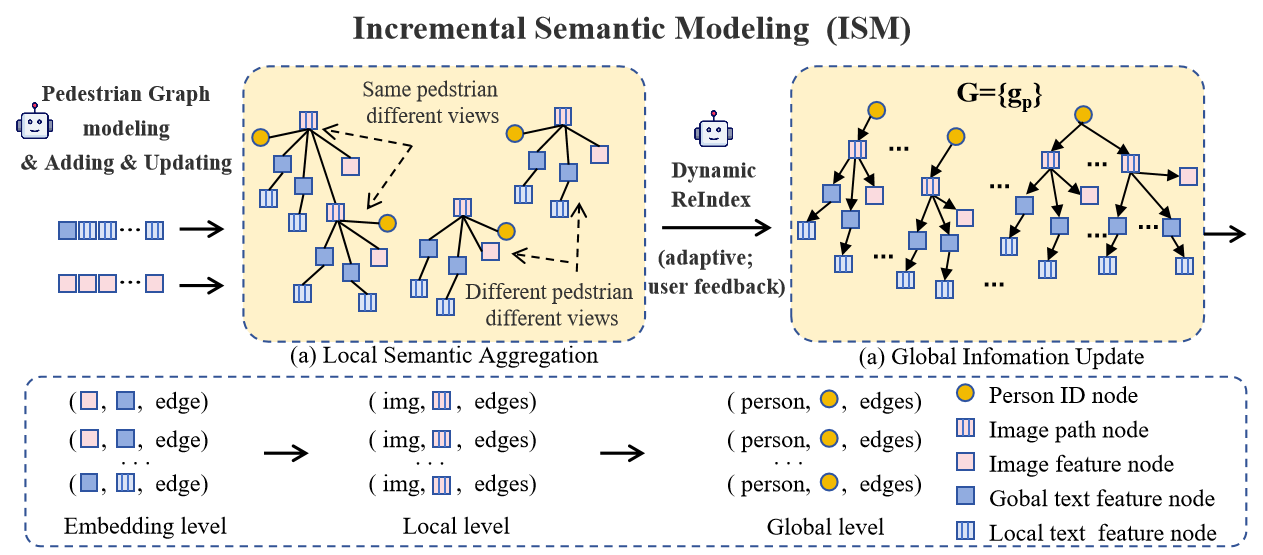}
	\caption{The proposed Incremental Semantic Mining (ISM) module.}
	\label{fig4}
\end{figure}

To address this limitation, we propose the Incremental Semantic Mining (ISM) module(Fig.~\ref{fig4}). ISM dynamically integrates reliable semantic cues from knowledge graphs, user feedback, and continuously updated data, thereby constructing a comprehensive representation of pedestrian instances. 
This mechanism effectively alleviates the limitations of single-round retrieval by modeling incremental semantics, enabling robust matching under diverse viewpoints and evolving descriptions.

The core of ISM lies in establishing a knowledge-graph-based incremental semantic representation by leveraging multi-round user feedback and expanded queries. The overall process contains three stages: \textit{graph construction}, \textit{local information aggregation}, and \textit{global index updating}.

\textbf{Graph Construction:}  Given an optimized pedestrian image $I_{opt}$ and its structured textual descriptions (original $Y_{ori}$ and generated $Y_{ren}$), the visual encoder extracts multimodal features $v_{img}$, 
while the textual input is tokenized into semantic nodes $v_{c}$. The resulting semantic state set $S=\{s_{h}, s_{u}, s_{l}, s_{a}\}$ corresponds to structured attributes of head, upper body, lower body, and accessories. 
To guarantee semantic consistency across multiple images, a unique identity $vid=Hash(I_{opt})$ is assigned. A single-image multi-relational graph $G_{p}^{(0)}$ is then constructed as:
\begin{equation}
\varepsilon_{p}^{(0)} = \{(v_{img}, v_{c}^{describe}, v_{id}), (v_{d}^{contain}, v_{c}), (v_{inc}^{belong}, v_{id})\},
\end{equation}
where image nodes, semantic nodes, and identity nodes are connected through descriptive and ownership relations.

\textbf{Local Information Aggregation:}  
For images corresponding to the same pedestrian identity (Fig.~\ref{fig4}(a)), local subgraphs are aggregated as:
\begin{equation}
G_{p}^{(1)} = \bigcup G_{p}^{(0)}, \quad I_{p} = \{id \mid v_{id}^{p}\}.
\end{equation}
To enhance semantic connectivity across related images, weak semantic edges are introduced between nodes with high similarity:
\begin{equation}
\varepsilon_{p}^{(1)} = \varepsilon_{p}^{(0)} \cup \{(v_{c}^{i}, v_{c}^{j}, v_{c}^{same}) \mid Sim(v_{c}^{i}, v_{c}^{j}) > \theta\},
\end{equation}
where $Sim(\cdot)$ denotes cosine similarity and $\theta$ is empirically set to 0.5.

\textbf{Global Index Updating:}  
After aggregating all local graphs for each identity (Fig.~\ref{fig4}(b)), a global semantic graph is obtained:
\begin{equation}
G = \bigcup g_{p} \mid p \in P.
\end{equation}
During query expansion with user feedback, new entities $e_{new}$ are attached to their corresponding identity nodes, with semantic relations defined as:
\begin{equation}
g_{p}^{r} = \{(e_{new}, r, e_{cb}) \mid r \in R_{val}\},
\end{equation}
where $R_{val}$ denotes the set of valid semantic relations.

Finally, all updated subgraphs are merged into the unified global graph $\hat{G}=\bigcup g_{p}$, which encodes consistent and expandable pedestrian semantics. 
This structure ensures that multi-turn retrieval incorporates dynamic graph updates, leading to robust and reliable semantic modeling across diverse user interactions. 
\subsection{Query-Aware Hierarchical Retrieval}\label{QHR}
To address the challenges of diversity and novelty in open-world pedestrian retrieval, this work introduces a Query-Aware Hierarchical Retrieval (QHR) module built upon the knowledge graph (KG) and multimodal large model reasoning. 
QHR dynamically expands queries and ensures effective alignment with candidate instances throughout the retrieval process. 

The module performs hierarchical vision-language matching by separately modeling textual and visual components of the query.
For the textual component, QHR computes the fusion similarity between text nodes (\textit{txtd}) and image nodes (\textit{img}), followed by attention-based ranking according to authority and similarity. 
For the visual component, QHR measures the similarity between the query image and candidate image nodes. The retrieved candidates are then re-ranked using graph-based reasoning, ensuring consistency between text and image semantics.

Given a query $Q$, we first extract its visual embedding $img_{q}$, textual embedding $txt_{q}$, and a set of contextual semantic nodes $\{sctxt_{q}^{1}, \ldots, sctxt_{q}^{M}\}$. 
These nodes are inserted into the knowledge graph as pseudo-query nodes to enrich the historical context. The initial similarity score is computed as follows:
\begin{equation}
Sim_{tgt}(Q) = \gamma \cdot Sim(Q, img) + (1-\gamma) \cdot Sim(Q, txt),
\end{equation}
where $\gamma$ is a modality-fusion weight controlling the contribution of text and image similarities.  

For each candidate instance $p$, we compute its text similarity $S_{txt}(p)$ and image similarity $S_{img}(p)$. The combined initial similarity score is defined as:
\begin{equation}
S_{init}(p) = \alpha \cdot S_{txt}(p) + \beta \cdot S_{img}(p), \quad \alpha + \beta = 1,
\end{equation}
where $\alpha$ and $\beta$ are balancing weights for text and image scores, respectively. 

After selecting the top-$N$ candidate instances, fine-grained semantic matching is performed at the node level. 
Each semantic node is compared with the query semantic nodes, and the hierarchical similarity score is defined as:
\begin{equation}
S_{sctxt}(p) = \sum_{k} w_{k} \cdot Sim(sctxt_{q}^{k}, sctxt_{p}^{k}),
\end{equation}
where $w_{k}$ denotes the importance weight of each semantic component.

Through this hierarchical scoring mechanism, QHR effectively integrates textual and visual semantics while dynamically incorporating query expansion. 
This ensures robust retrieval performance in open-world scenarios with evolving queries.
\section{Experiments}\label{sec4}
\subsection{Experimental Settings}
\textbf{Datasets.}
We evaluate \textit{FitPro} on five benchmarks that cover both \textbf{closed-world} and \textbf{open-world} retrieval settings. 
The \textbf{closed-world} setting is represented by three widely used cropped-image datasets:  
\textbf{CUHK-PEDES}~\cite{CUHK} with 40,206 images of 13,003 identities (two descriptions per image);  
\textbf{RSTPReid}~\cite{RSTP} containing 20,505 images of 4,101 identities (two descriptions per image);  
and \textbf{ICFG-PEDES}~\cite{ICFG} comprising 54,522 images of 4,102 identities, each annotated with one description.
For the \textbf{open-world} setting, we further evaluate on two full-scene benchmarks:  
The \textbf{CUHK-SYSU-TBPS}~\cite{SDRPN} training set contains 11,206 scene images with 15,080 bounding boxes from 5,532 identities, where each box has two textual descriptions, while the query set has 2,900 boxes with one description each. 
The \textbf{PRW-TBPS}~\cite{SDRPN}, the training set includes 5,704 images with 14,897 bounding boxes covering 483 identities (one description per box), and the query set consists of 2,056 boxes annotated with two descriptions. 
All experiments follow a \textbf{zero-shot} protocol on the official test splits to ensure fair and consistent comparison.

\textbf{Architecture and Inference Protocol.}
We employ \textbf{ALBEF}~\cite{MARS} as the default vision-language backbone. 
Image enhancement in FCD is performed using a \textbf{U-Net}~\cite{ID-Blau} with lossless upsampling (denoising step size = 20), and text decoding is implemented with \textbf{LLaVA}~\cite{llava}.  
To validate the generality of our hallucination suppression strategy, we also evaluate \textbf{BLIP}~\cite{BLIP}, \textbf{Qwen}~\cite{qwen}, and \textbf{Mini-GPT4}~\cite{MiniGPT4}.  
All results are obtained under zero-shot inference without training or fine-tuning on target datasets.  
To prevent any potential data leakage, it is worth noting that only the pedestrian identities represented by the hash-based pseudo-IDs described in Section~\ref{QHR} are updated, 
rather than the ground-truth IDs used in the computation of evaluation metrics.
Furthermore, to better simulate and evaluate real-world interactive retrieval scenarios,each interaction reveals only a single correct answer instead of exposing all possible matches at once.

\textbf{Evaluation Metrics.}
We use \textbf{Rank-K} (K=1, 5, 10) to evaluate overall recognition capability and retrieval accuracy.  
Rank-K imposes stricter requirements on returned ranking positions than the commonly used Hits@K in traditional interactive systems,  
thus providing a more reliable measure of performance in open-world scenarios.  
To assess ranking quality over all retrieved results, we use \textbf{Mean Average Precision (mAP)}, 
as it more comprehensively reflects the system's sensitivity to mis-ranking and is suitable for evaluating retrieval effectiveness under varying recall levels.  
Since traditional no-reference image quality metrics (e.g., NIQE~\cite{NIQE}, BRISQUE~\cite{BRISQUE}) cannot effectively capture the contribution of image details to downstream retrieval,  
we adopt the \textbf{POPE} framework~\cite{Evaluating} to evaluate the impact of our denoising and deblurring strategies on hallucination suppression.  
POPE measures hallucination indirectly through the \emph{Accuracy} and \emph{Precision} of generated descriptions,  
making it more appropriate for structured description tasks where both fine-grained detail preservation and semantic correctness are critical.

\begin{table}[!t]
\caption{Performance Comparison (\%) of FitPro and State-of-the-art Methods on CUHK-PEDES Dataset}
\label{tab1}
\centering
\begin{tabular}{|l|c|c|c|c|c|}
\hline
\textbf{Methods} & \textbf{Ref} & \textbf{mAP} & \textbf{Rank-1} & \textbf{Rank-5} & \textbf{Rank-10} \\
\hline
\multicolumn{6}{|l|}{\textbf{Non-Interactive}} \\
\hline
IRRA\cite{IRRA} & CVPR$^{23}$ & 66.13 & 73.38 & 89.93 & 93.71 \\
APTM\cite{APTM} & MM$^{23}$ & 66.91 & 76.53 & 90.04 & 94.15 \\
RDE\cite{RDE} & CVPR$^{24}$ & 67.56 & 75.94 & 90.14 & 94.12 \\
RaSA\cite{RASA} & IJCAI$^{23}$ & 69.38 & 76.51 & 90.29 & 94.25 \\
MARS\cite{MARS} & Arxiv$^{24}$ & 71.41 & 77.62 & 90.63 & 94.27 \\
\textbf{FitPro} & -- & \textbf{71.89} & \textbf{78.40} & 90.60 & 94.19 \\
\hline
\multicolumn{6}{|l|}{\textbf{Interactive}} \\
\hline
CIM\cite{CIM} (r=6) & MM$^{24}$ & -- & 78.06 & 91.71 & 95.69 \\
FitPro (r=6) & -- & \textbf{88.26} & \textbf{92.15} & \textbf{98.81} & \textbf{99.53} \\
ChatReID\cite{ChatReID} & Arxiv$^{25}$ & 80.1 & 83.8 & -- & -- \\
\textbf{FitPro} (r>6) & -- & \textbf{88.72} & \textbf{95.35} & \textbf{97.67} & \textbf{99.89} \\
\hline
\end{tabular}
\end{table}

\begin{table}[!t]
\caption{Performance Comparison (\%) of FitPro and State-of-the-art Methods on RSTPReid Dataset}
\label{tab2}
\centering
\begin{tabular}{|l|c|c|c|c|c|}
\hline
\textbf{Methods} & \textbf{Ref} & \textbf{mAP} & \textbf{Rank-1} & \textbf{Rank-5} & \textbf{Rank-10} \\
\hline
\multicolumn{6}{|l|}{\textbf{Non-Interactive}} \\
\hline
IRRA\cite{IRRA} & CVPR$^{23}$ & 47.17 & 60.20 & 81.30 & 88.20 \\
APTM\cite{APTM} & MM$^{23}$ & 52.56 & 67.50 & 85.70 & 91.45 \\
RDE\cite{RDE} & CVPR$^{24}$ & 50.88 & 65.35 & 85.95 & 90.90 \\
RaSA\cite{RASA} & IJCAI$^{23}$ & 52.31 & 66.90 & 86.00 & 91.35 \\
MARS\cite{MARS} & Arxiv$^{24}$ & 52.91 & 67.55 & 86.05 & 91.40 \\
\textbf{FitPro} & -- & \textbf{53.54} & \textbf{68.25} & \textbf{86.70} & \textbf{91.50} \\
\hline
\multicolumn{6}{|l|}{\textbf{Interactive}} \\
\hline
CIM\cite{CIM} (r=6) & MM$^{24}$ & -- & 75.05 & 88.95 & 94.15 \\
FitPro (r=6) & -- & \textbf{69.80} & \textbf{84.80} & \textbf{95.50} & -- \\
ChatReID\cite{ChatReID} & Arxiv$^{25}$ & 73.1 & 75.0 & -- & -- \\
\textbf{FitPro} (r>6) & -- & \textbf{73.28} & \textbf{87.50} & \textbf{96.50} & \textbf{98.50} \\
\hline
\end{tabular}
\end{table}

\begin{table}[!t]
\caption{Performance Comparison (\%) of FitPro and State-of-the-art Methods on ICFG-PEDES Dataset}
\label{tab3}
\centering
\begin{tabular}{|l|c|c|c|c|c|}
\hline
\textbf{Methods} & \textbf{Ref} & \textbf{mAP} & \textbf{Rank-1} & \textbf{Rank-5} & \textbf{Rank-10} \\
\hline
\multicolumn{6}{|l|}{\textbf{Non-Interactive}} \\
\hline
IRRA\cite{IRRA} & CVPR$^{23}$ & 38.06 & 63.46 & 80.25 & 85.82 \\ 
APTM\cite{APTM} & MM$^{23}$ & 41.22 & 68.51 & 82.99 & 87.56 \\ 
RDE\cite{RDE} & CVPR$^{24}$ & 40.06 & 67.68 & 82.47 & 87.36 \\ 
RaSA\cite{RASA} & IJCAI$^{23}$ & 41.29 & 65.28 & 80.40 & 85.12 \\ 
MARS\cite{MARS} & Arxiv$^{24}$ & 44.93 & 67.60 & 81.47 & 85.79 \\ 
\textbf{FitPro} & -- & \textbf{45.83} & \textbf{68.50} & \textbf{82.80} & \textbf{85.80} \\ 
\hline
\multicolumn{6}{|l|}{\textbf{Interactive}} \\
\hline
CIM\cite{CIM} (r=6) & MM$^{24}$ & -- & 71.92 & 86.08 & 90.55 \\ 
FitPro (r=6) & -- & \textbf{65.47} & \textbf{90.50} & \textbf{97.90} & \textbf{99.20} \\ 
ChatReID\cite{ChatReID} & Arxiv$^{25}$ & 70.5 & 72.9 & -- & -- \\ 
\textbf{FitPro} (r>6) & -- & \textbf{69.68} & \textbf{91.60} & \textbf{98.60} & \textbf{99.50} \\ 
\hline
\end{tabular}
\end{table}
\subsection{Performance Comparison on Cropped Images}
Tables~\ref{tab1}-\ref{tab3} compare \textit{FitPro} with state-of-the-art methods on CUHK-PEDES~\cite{CUHK}, RSTPReid~\cite{RSTP}, and ICFG-PEDES~\cite{ICFG} under both conventional and interactive settings.
In the \textbf{non-interactive setting}, \textit{FitPro} consistently achieves the best results on all three datasets. 
It obtains 71.89\%/78.40\% (mAP/Rank-1) on CUHK-PEDES, 53.54\%/68.25\% on RSTPReid, and 45.83\%/68.50\% on ICFG-PEDES, 
demonstrating the strong modeling capability and robustness of \textit{FitPro} for the language-image matching task under non-interactive retrieval settings.

In the \textbf{interactive setting}, approaches that directly apply BERT-based or encoder-based retrieval models often degrade due to word-order variations and inconsistent query rewriting. 
In contrast, \textit{FitPro} achieves stable improvements across interaction rounds.  
On CUHK-PEDES, performance reaches 88.26\%/92.15\% (mAP/Rank-1) at $r=6$ and further improves to 88.72\%/95.35\% when $r>6$. 
On RSTPReid, Rank-1 increases from 84.50\% at $r=6$ to 87.50\% at $r>6$. 
On ICFG-PEDES, Rank-1 improves from 90.50\% to 91.60\%. 
\textit{FitPro} also achieves higher Rank-5 and Rank-10 scores than CIM\cite{CIM} and ChatReID\cite{ChatReID} across all datasets.
Overall, the results validate the effectiveness and scalability of \textit{FitPro} in both non-interactive and interactive retrieval scenarios.
Its gains mainly come from:  
(i) the reliable prompt expansion strategy that enhances text-visual alignment, and  
(ii) the efficient hierarchical retrieval strategy that captures key semantics in expanded queries.  
The inherent text-to-image matching ability and open-domain understanding of modern MLLMs further strengthen its generalization across datasets.
\subsection{Performance Comparison on Scene Images}
Table~\ref{tab4} compares \textit{FitPro} with SDRPN~\cite{SDRPN}, MACA~\cite{MACA}, and UPD~\cite{MACA} on the open-domain CUHK-SYSU-TBPS and PRW-TBPS datasets.  
\textit{FitPro} achieves the highest accuracy on all metrics, reaching 71.99\% mAP on CUHK-SYSU-TBPS and 60.31\% on PRW-TBPS, representing clear improvements over existing scene-level TPR methods.  
These results confirm the strong effectiveness of \textit{FitPro} for interactive text-based person retrieval in open-world scenarios.
\begin{table*}[!t]
\caption{Performance Comparison (\%) of FitPro and State-of-the-art Methods on Two Open-Scene Datasets}
\label{tab4}
\centering
\begin{tabular}{|l|c|c|c|c|c|c|c|c|}
\hline
\multirow{2}{*}{\textbf{Methods}} & \multicolumn{4}{c|}{\textbf{CUHK-SYSU-TBPS}} & \multicolumn{4}{c|}{\textbf{PRW-TBPS}} \\ 
\cline{2-9}
 & \textbf{mAP} & \textbf{Rank-1} & \textbf{Rank-5} & \textbf{Rank-10} & \textbf{mAP} & \textbf{Rank-1} & \textbf{Rank-5} & \textbf{Rank-10} \\ 
\hline
SDRPN\cite{SDRPN} & 50.36 & 49.34 & 74.48 & 82.14 & 11.93 & 21.63 & 42.54 & 52.99 \\ 
MACA\cite{MACA} & 57.77 & 52.03 & 76.71 & 83.79 & 18.18 & 33.25 & 52.87 & 61.93 \\ 
UPD\cite{UPD} & 57.43 & 57.95 & 77.36 & 84.83 & 17.56 & 37.54 & 53.55 & 62.68 \\ 
\textbf{FitPro} & \textbf{71.99} & \textbf{82.35} & \textbf{92.86} & \textbf{95.38} & \textbf{60.31} & \textbf{77.85} & \textbf{91.77} & \textbf{95.56} \\ 
\hline
\end{tabular}
\end{table*}

\begin{table*}[!t]
\caption{Performance Comparison (\%) of Different Component Combinations on Three Public Datasets for Interactive Retrieval}
\label{tab5}
\centering
\begin{tabular}{|c|l|c|c|c|c|c|c|}
\hline
\textbf{No.} & \textbf{Components} & \multicolumn{2}{c|}{\textbf{CUHK-PEDES}} & \multicolumn{2}{c|}{\textbf{RSTPReid}} & \multicolumn{2}{c|}{\textbf{ICFG-PEDES}} \\ 
\cline{3-8}
 &  & \textbf{mAP} & \textbf{Rank-1} & \textbf{mAP} & \textbf{Rank-1} & \textbf{mAP} & \textbf{Rank-1} \\ 
\hline
1 & Baseline & 52.4 & 62.6 & 40.9 & 52.8 & 38.7 & 50.2 \\ 
2 & + FCD & 58.3 & 68.7 & 45.6 & 58.3 & 42.9 & 55.1 \\ 
3 & + ISM & 63.4 & 73.2 & 48.2 & 61.7 & 46.3 & 59.8 \\ 
4 & + QHR ($r < 3$) & 67.8 & 76.5 & 50.1 & 64.2 & 49.7 & 63.4 \\ 
5 & FCD+ISM+QHR ($r \geq 3$) & \textbf{71.9} & \textbf{78.4} & \textbf{53.5} & \textbf{68.3} & \textbf{52.8} & \textbf{66.7} \\ 
\hline
\end{tabular}
\end{table*}

\begin{table*}[!t]
\caption{Performance Comparison (\%) of Different Multimodal Large Language Models in FCD for Interactive Retrieval}
\label{tab6}
\centering
\begin{tabular}{|l|c|c|c|c|c|c|c|c|}
\hline
\textbf{MLLMs} & \multicolumn{4}{c|}{\textbf{CUHK-PEDES}} & \multicolumn{4}{c|}{\textbf{RSTPReid}} \\ 
\cline{2-9}
 & \textbf{mAP} & \textbf{Rank-1} & \textbf{Rank-5} & \textbf{Rank-10} & \textbf{mAP} & \textbf{Rank-1} & \textbf{Rank-5} & \textbf{Rank-10} \\ 
\hline
BLIP\cite{BLIP} & 52.4 & 62.6 & 64.9 & 70.3 & 40.9 & 52.8 & 63.2 & 75.5 \\ 
Qwen\cite{qwen} & 63.4 & 73.2 & 88.7 & 92.6 & 44.7 & 62.5 & 75.9 & 84.3 \\ 
LLaVA\cite{llava} & \textbf{71.9} & \textbf{78.4} & \textbf{90.6} & \textbf{94.3} & \textbf{53.5} & \textbf{68.3} & \textbf{86.7} & \textbf{91.5} \\ 
\hline
\end{tabular}
\end{table*}
\subsection{Ablation studies}
To evaluate the contribution of each module, ablation studies are conducted on CUHK-PEDES, RSTPReid, and ICFG-PEDES, with results summarized in Table~\ref{tab5}. 
In our setup, $r<3$ denotes using only textual descriptions of the retrieved person, while $r \geq 3$ additionally incorporates information from the person's entity graph. 
"w/o ISM" removes the Incremental Semantic Mining module, where structured semantic texts are not extracted via semantic templates. 
Introducing the FCD module yields slight improvements across all datasets. On CUHK-PEDES, mAP/Rank-1 increase by 0.52\%/0.81\%; on RSTPReid, by 0.62\%/0.70\%; and on ICFG-PEDES, by 0.90\%/0.90\%. 
These results show that enhancing data quality and reducing hallucinations in MLLMs benefits retrieval, though the gains remain modest.
Incorporating ISM leads to a substantial boost, with average improvements of 6.43\% in mAP and 7.94\% in Rank-1. 
ICFG-PEDES reaches 54.43\% mAP and 78.80\% Rank-1, validating the importance of constructing global structured representations for diverse attributes of the same person. 
Adding the QHR strategy further improves performance. 
When $r<3$, scores reach 83.21\%/87.99\% on CUHK-PEDES, 63.33\%/77.52\% on RSTPReid, and 58.96\%/84.59\% on ICFG-PEDES, 
indicating that leveraging global structured representations of pedestrian entities during query expansion enhances the model's capability in understanding and generating responses to complex queries.  
When $r \geq 3$, performance peaks: 88.72\%/95.35\% on CUHK-PEDES, 73.28\%/87.50\% on RSTPReid, and 69.68\%/91.60\% on ICFG-PEDES. 
These results provide strong evidence that deeply mining semantic associations in the query evolution process can substantially improve retrieval accuracy and the handling of complex queries.
\begin{figure}[!t]
  \centering
  \includegraphics[width=\columnwidth]{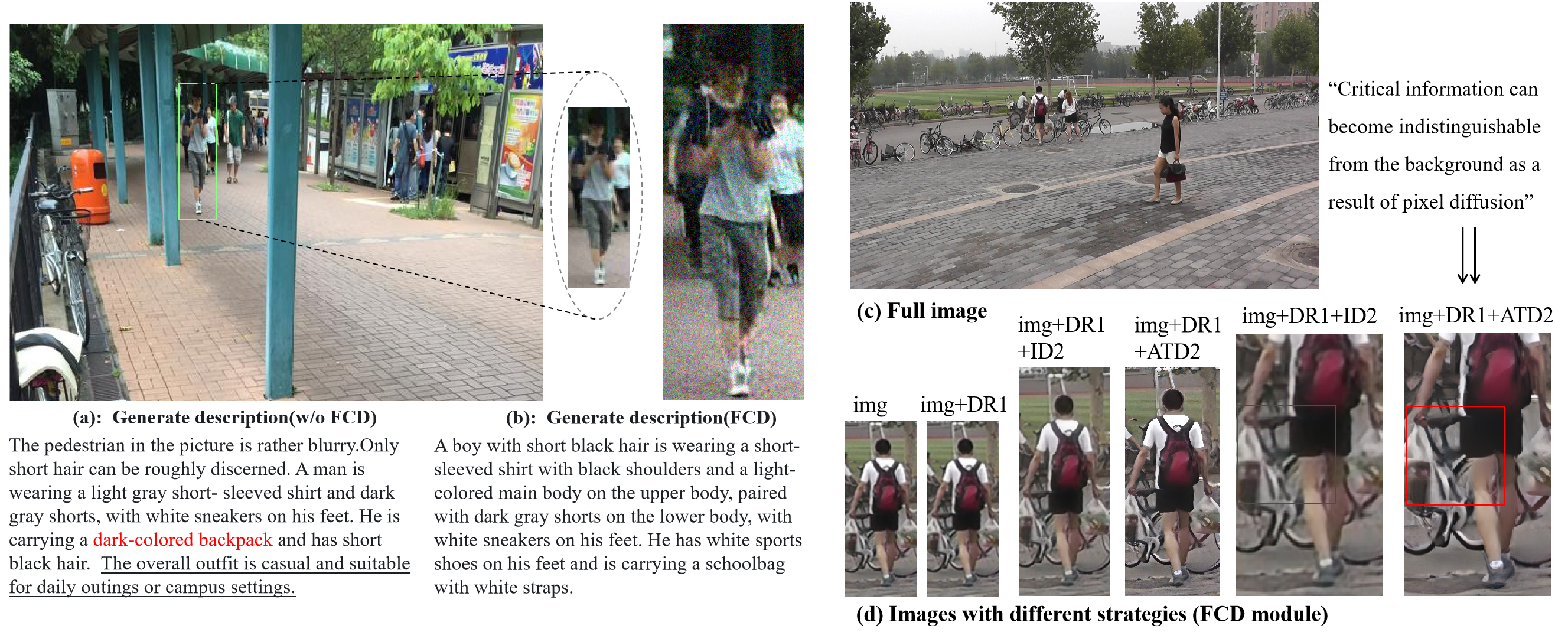}
  \caption{Visualization of Different Image Denoising and Lossless Upscaling Strategies in the FCD Module. The restored details mainly enhance pedestrian-related textures (appearance, clothing, and accessories) rather than introducing noise.}
  \label{fig5}
\end{figure}
\begin{table*}[!t]
\caption{Performance Comparison (\%) of Different Methods for Mitigating Visual Hallucinations in FCD}
\label{tab7}
\centering
\begin{tabular}{|l|c|c|c|c|c|c|c|c|}
\hline
\multirow{2}{*}{\textbf{Methods}} & \multicolumn{2}{c|}{\textbf{PRW-TBPS}} & \multicolumn{2}{c|}{\textbf{CUHK-SYSU-TBPS}} & \multicolumn{2}{c|}{\textbf{CUHK-PEDES}} & \multicolumn{2}{c|}{\textbf{RSTPReid}} \\ 
\cline{2-9}
 & \textbf{Accuracy} & \textbf{Precision} & \textbf{Accuracy} & \textbf{Precision} & \textbf{Accuracy} & \textbf{Precision} & \textbf{Accuracy} & \textbf{Precision} \\ 
\hline
Imgs & 66.34 & 94.83 & 68.51 & 90.58 & 66.11 & 88.55 & 65.30 & 83.10 \\ 
Imgs+DR1 & 66.45 & 95.45 & 68.58 & 90.75 & 66.11 & 88.77 & 65.30 & 83.16 \\ 
Imgs+DR1+ATD2 & 66.34 & 94.95 & 68.58 & 90.73 & 66.10 & 88.69 & 65.30 & 83.10 \\ 
Imgs+DR1+ID2 & \textbf{71.76} & 93.55 & 65.78 & \textbf{94.89} & \textbf{66.47} & \textbf{92.01} & \textbf{65.40} & \textbf{84.90} \\ 
\hline
\end{tabular}
\end{table*}
\subsection{FCD performance to Retrieval and Visualization}
As shown in Table~\ref{tab6}, different multimodal large language models yield varying performance in the FCD stage. 
LLava-based FitPro achieves the best overall results on both CUHK-PEDES and RSTPReid, 
reaching 71.9\% and 53.5\% in mAP and 78.4\% and 68.7\% in Rank-1, 
respectively—substantially higher than BLIP (52.4\%, 32.6\%) and Qwen (63.4\%, 47.8\%). 
These improvements indicate that LLava-based FCD provides more stable and fine-grained text-image alignment across datasets and evaluation metrics.

Table~\ref{tab7} compares several image enhancement strategies used in FCD for mitigating visual hallucinations. 
"Imgs" denotes using raw inputs, "Imgs+DR1" applies first-stage denoising. "Imgs+DR1+ATD2" further adds diffusion-based upscaling, and incorporates structure-aware detail preservation before upscaling.    
Raw images suffer from strong noise interference, degrading the quality of structured descriptions. 
Basic denoising (Imgs+DR1) yields modest gains, while diffusion-based enhancement (ATD2) provides only limited improvements. 
In contrast, the structure-aware ID2 strategy achieves the best performance across all datasets; on PRW-TBPS, accuracy rises to 71.76\%.  
Although precision slightly decreases due to low image resolution, ID2 demonstrates stronger robustness and generalization. 
Overall, the multi-stage enhancement in FCD—especially Imgs+DR1+ID2—effectively reduces hallucination and improves consistency in text generation.

\begin{figure*}[!t]
\centering
\subfloat[]{\includegraphics[width=3.5in]{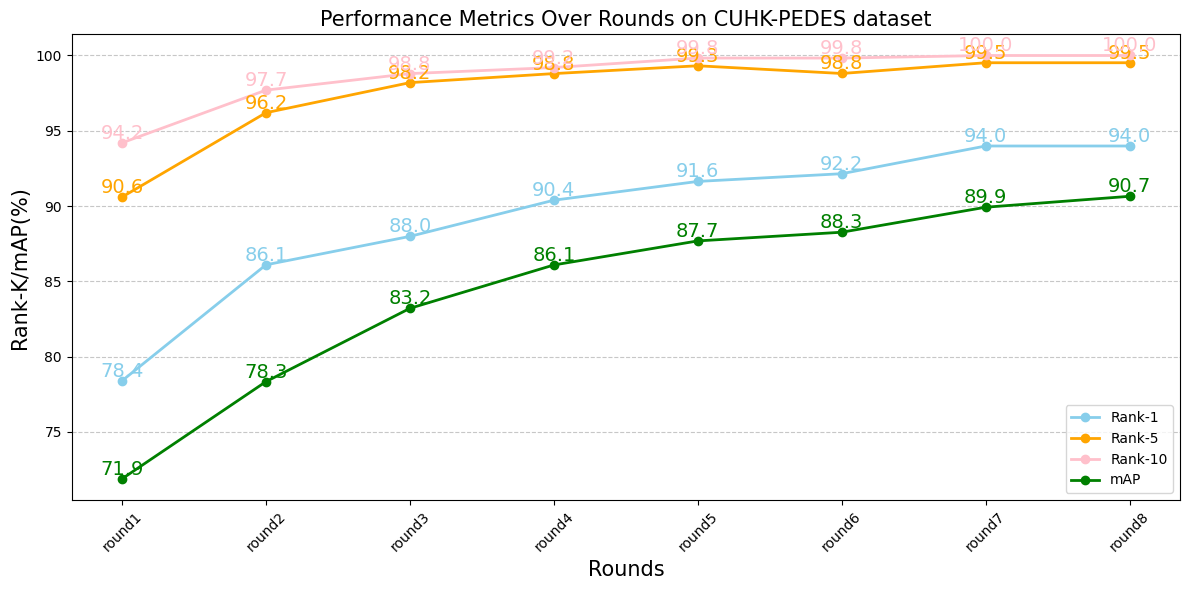}%
\label{fig6.1}}
\hfil
\subfloat[]{\includegraphics[width=3.5in]{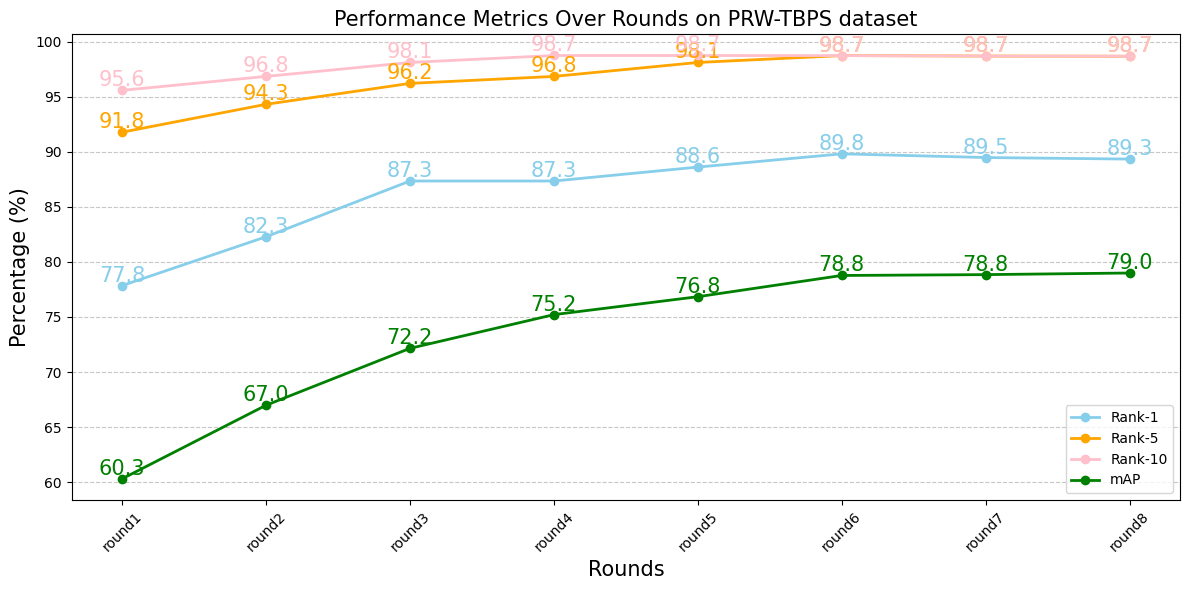}%
\label{fig6.2}}
\caption[Interactive Retrieval Performance Comparison]{Interactive retrieval performance comparison on conventional dataset CUHK-PEDES (left) and open-scene dataset PRW-TBPS (right). The visualization demonstrates the effectiveness of our method across different scenario types.}
\label{fig6}
\end{figure*}

Figure~\ref{fig6} further illustrates the effectiveness of FCD in alleviating visual hallucinations on the PRW-TBPS dataset. 
ATD2-based approaches often suffer from edge oversmoothing and structural blurring (red boxes), 
whereas ID2-based strategies preserve richer local structures and high-frequency details, 
thereby enhancing the expression of fine-grained attributes (e.g., clothing and accessories) in subsequent structured descriptions. 
This indicates that the structure-aware enhancement mechanism can better retain critical semantic features without sacrificing overall image consistency, improving downstream descriptive reliability.  

We also evaluate Mini-GPT4 on CUHK-SYSU-TBPS,as shown in Fig.~\ref{fig5}(a-b).
The results show that generating more detailed descriptions does not necessarily improve retrieval: 
Mini-GPT4 often introduces noisy or marginally relevant semantics (e.g., negation-based expressions such as "a white backpack without patterns"),  
which are sensitive to viewpoint changes and may lead to false negatives.
As red text in Fig.~\ref{fig5}(c) illustrates, descriptions generated from blurred images tend to contain inaccurate or marginally useful details, aggravating visual hallucination effects and degrading overall text quality.
By contrast, the proposed LLava-based method, combined with visual structural templates, produces cleaner and more discriminative descriptions (Fig.~\ref{fig5}(d)), enhancing both specificity and usability. 
Therefore, the proposed framework effectively mitigates retrieval errors caused by image blurring and visual hallucinations through its feature-contrastive decoding strategy, 
ultimately improving the accuracy and stability of structured pedestrian representations. 
\subsection{Different interactive turns with QHR to Retrieval results}
As shown in Fig.~\ref{fig6}, we further investigate the effect of iterative interaction across two settings—the conventional setting (CUHK-PEDES) and the open-domain setting (PRW-TBPS). 
The results show that both Rank-$k$ ($k=1,5,10$) and mAP steadily improves as the number of iterations increases, though gains become marginal after five rounds, with Rank-5 and Rank-10 nearly saturated. 
This suggests that the model can already capture the main discriminative local cues (e.g., head, upper body, lower body, accessories) through FCD, while ISM provides structured global representations of the same identity.  
Although mAP starts relatively low, it shows a large improvement, with an average increase of about 18\% across the two datasets. 
Nonetheless, Rank-1 and mAP still exhibit room for improvement after the eighth iteration. 
This remaining gap mainly stems from insufficient modeling of fine-grained attributes (e.g., "white sneakers with black shoelaces"),
while variations in image resolution and viewpoint further introduce instability in capturing subtle local details and limit overall retrieval accuracy.
\subsection{Discussion}
\textbf{Effectiveness of interactive retrieval.} FitPro achieves notable gains in interactive settings. 
The ISM module incrementally incorporates user feedback into structured entity graphs, 
progressively refining person representations and reducing semantic ambiguity, 
which explains the consistent improvements in Rank-1 and mAP across turns.  

\textbf{Adaptability in open-scene datasets.}On CUHK-SYSU-TBPS and PRW-TBPS, FitPro yields higher accuracy under complex backgrounds and viewpoint changes. 
FCD enhances robustness to noise through structure-aware visual representations, whereas QHR adjusts retrieval based on query completeness. 
Their combination maintains both precision and recall, outperforming non-interactive baselines in unconstrained environments.

\textbf{Synergy of modules.} Ablation results show that all modules contribute incrementally, while their joint use offers the largest gains. 
FCD provides stable alignment, ISM models evolving semantics, and QHR enables adaptive retrieval. 
Together, they form a complementary system that improves robustness, adaptability, and scalability in open-world TPR.
\section{Conclusion}\label{sec5}
In this paper, we propose \textit{FitPro}, an interactive zero-shot text-based person retrieval framework tailored for open-world scenarios. 
To address the challenges of diverse visual conditions and dynamic query refinement, FitPro integrates three core modules:  
(i) Feature-Contrastive Decoding (FCD) for robust structured representations; 
(ii) Incremental Semantic Mining (ISM) for modeling viewpoint and description variations; 
(iii) Query-aware Hierarchical Retrieval (QHR) for adaptive multimodal retrieval. 
Experimental results demonstrate that \textit{FitPro} achieves superior performance across multiple datasets, providing effective support for practical deployment in real-world applications.  
Beyond its immediate contributions, this work provides theoretical insights into the mutual reinforcement between interactive modeling and zero-shot generalization, 
and demonstrates practical feasibility for deployment in dynamic, human-centered surveillance systems.

\vspace{11pt}

\vfill

\end{document}